\documentclass{article}

\usepackage{arxiv}

\usepackage[utf8]{inputenc} 
\usepackage[T1]{fontenc}    
\usepackage{hyperref}       
\usepackage{url}            
\usepackage{booktabs}       
\usepackage{amsfonts}       
\usepackage{nicefrac}       
\usepackage{microtype}      
\usepackage{amsmath}
\usepackage{cleveref}       
\usepackage{lipsum}         
\usepackage{graphicx}
\usepackage{doi}

\usepackage{subcaption}
\usepackage{algorithm}
\usepackage{algpseudocode}
\makeatletter
\algrenewcommand\ALG@beginalgorithmic{\ttfamily}
\makeatother
\usepackage{wrapfig}

\usepackage[backend=biber, sorting=none, style=numeric]{biblatex}
\title{Rational Linkages: From Poses to 3D-printed Prototypes}

\newif\ifuniqueAffiliation

\ifuniqueAffiliation 
\author{ \href{https://orcid.org/0000-0002-7398-7825}{\includegraphics[scale=0.06]{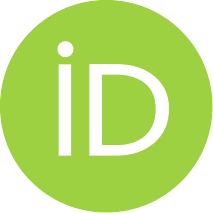}\hspace{1mm}Daniel Huczala}\\
	Unit of Geometry and Surveying\\
	University of Innsbruck\\
	Innsbruck, Austria \\
	\texttt{daniel.huczala@uibk.ac.at} \\
	\And 
	\href{https://orcid.org/0000-0000-0000-0000}{\includegraphics[scale=0.06]{orcid.pdf}\hspace{1mm}Johannes Siegele} \\
	Johann Radon Institute for Computational \\ and Applied Mathematics (RICAM) \\
	Austrian Academy of Sciences \\
	Linz, Austria \\
	\And 
	\href{https://orcid.org/0000-0000-0000-0000}{\includegraphics[scale=0.06]{orcid.pdf}\hspace{1mm}Daren A. Thimm} \\
	Unit of Geometry and Surveying\\
	University of Innsbruck\\
	Innsbruck, Austria \\
        \And 
	\href{https://orcid.org/0000-0000-0000-0000}{\includegraphics[scale=0.06]{orcid.pdf}\hspace{1mm}Martin Pfurner} \\
	Unit of Geometry and Surveying\\
	University of Innsbruck\\
	Innsbruck, Austria \\
        \And 
	\href{https://orcid.org/0000-0000-0000-0000}{\includegraphics[scale=0.06]{orcid.pdf}\hspace{1mm}Hans-Peter Schr\"ocker} \\
	Unit of Geometry and Surveying\\
	University of Innsbruck\\
	Innsbruck, Austria \\
}
\else
\usepackage{authblk}

\newbox{\orcid}\sbox{\orcid}{\includegraphics[scale=0.06]{orcid.pdf}} 
\author[1]{%
	\href{https://orcid.org/0000-0002-7398-7825}{\usebox{\orcid}\hspace{1mm}Daniel Huczala\thanks{\texttt{daniel.huczala@uibk.ac.at}}}%
}
\author[2]{%
	\href{https://orcid.org/0000-0002-8790-0081}{\usebox{\orcid}\hspace{1mm}Johannes Siegele}%
}
\author[1]{%
	\href{https://orcid.org/0009-0003-6478-1782}{\usebox{\orcid}\hspace{1mm}Daren A. Thimm}%
}
\author[1]{%
	\href{https://orcid.org/0000-0003-1988-2202}{\usebox{\orcid}\hspace{1mm}Martin Pfurner}%
}
\author[1]{%
	\href{https://orcid.org/0000-0003-2601-6695}{\usebox{\orcid}\hspace{1mm}Hans-Peter Schr\"ocker}%
}
\affil[1]{Unit of Geometry and Surveying, University of Innsbruck, Innsbruck, Austria}
\affil[2]{Johann Radon Institute for Computational and Applied Mathematics (RICAM), Austrian Academy of Sciences, Linz, Austria}
\fi

\renewcommand{\shorttitle}{Rational Linkages: From Poses to 3D-printed Prototypes}

\hypersetup{
pdftitle={A template for the arxiv style},
pdfsubject={q-bio.NC, q-bio.QM},
pdfauthor={David S.~Hippocampus, Elias D.~Striatum},
pdfkeywords={First keyword, Second keyword, More},
}

\begin{document}

\maketitle

\begin{abstract}
In this paper, a set of tools is introduced that simplifies the synthesis and rapid-prototyping of single-loop rational kinematic chains. It allows the user to perform rational motion interpolation of up to four given poses and yields the design parameters of a linkage that can execute this motion. The package also provides a visualization of the output and performs a self-collision analysis with the possibility to adapt the design parameters. The results can be imported into CAD-systems for fast 3D printing.
\end{abstract}

\keywords{rational linkages \and single-loop mechanisms \and robot design \and rapid prototyping}

\section{Introduction}
Single-loop N-bar linkages are closed kinematic chains that connect a number of rigid bars via revolute or prismatic joints in one loop. Over-constrained mechanisms are a special class of such linkages. In this class, the Grübler-Kutzbach-Chebyshev formula, computing the degrees-of-freedom (DoF) of the mechanism, fails and the causes of this failure are special geometric properties of the design parameters of the mechanism. Because of the low number of DoFs, single-loop linkages can be synthesized as simple single-purpose devices for specific tasks, such as pick-and-place operations. This can be advantageous compared to serial robots, because they can perform complicated motions following space-curves of higher degree with a low number of active joints. Up to now, a major problem for the industrial application of these mechanisms is self-intersections \cite{Li2020}, as discussed later.

Custom mechanisms designed for specific tasks are a wide research topic, with many studies conducted in fields such as robot manipulation \cite{Yihun2014, Dogra2022, Huczala2022} or locomotion \cite{Xi2018, Zhang2019}, whereby the design parameters are mostly obtained by numerical optimization. The tools presented in this paper, however, rely on algebraic methods. The problem of designing a spatial linkage analytically can be divided into the following steps: Pose interpolation, synthesis of the kinematic structure, and design of its physical realization, which also involves a collision analysis and CAD modeling to prepare 3D printing.

The main objective of this research is to bring single-loop linkages closer to practical applications by providing an open-source Python package that can be easily used, extended, and maintained. The methodologies that have been implemented and their handling are described in the following sections. So far, the focus has been placed on 1-DoF single-loop linkages with four or six revolute (4R or 6R) joints that perform rational motions. These linkages benefit from the rational mathematical representation of the motion, which allows factorization and parametrization of relative motions.

The prime tasks in designing a custom mechanism or manipulator is the determination of the poses, which should be achieved by the end-effector tool of the device. The presented package implements for this purpose a method introduced by Hegedüs et al. \cite{Hegeds2015}, which interpolates four poses using a rational curve in the special Euclidean displacement group SE(3). When a rational motion curve on SE(3) is obtained, rational motion factorization introduced by Hegedüs, Schicho, and Schröcker \cite{Hegeds2012, Hegeds2013} is applied. The factors of the rational motion curve relate to 1-DoF revolute or prismatic joints, which can be connected to form a linkage capable of performing the synthesized end-effector motion.


A further feature of the Rational Linkages package is the possibility to manually model the physical realization of the linkage and to generate design parameters that can be inserted into pre-prepared CAD models for immediate 3D printing.

The paper is organized as follows: Section 2 provides a short introduction into the mathematical background and in Section 3 it will be shown how this toolbox can take a user from the given poses to a 3D printed mechanism prototype in a few lines of code. The paper ends with a conclusion and a hint to a web page that provides more detailed information on the introduced Rational Linkage package.

\section{Mathematical Background}\label{sec:mathback}
This section is a very short introduction to the implemented mathematical methods and provides references to the works in which they were introduced or discussed.

\vspace{10pt}\noindent
\textbf{Pl{\"u}cker Coordinates, Dual Quaternions, and Rational Motions}
\vspace{5pt}

\noindent
A possibility to describe lines in 3D space using coordinates are the so called Pl{\"u}cker coordinates. These are six-dimensional vectors $\mathbf{l}=(g_0, g_1, g_2, g_3, g_4, g_5)$. The line's direction is $\mathbf{g}=(g_0,g_1,g_2)$, its moment vector $\overline{\mathbf{g}} = (g_3,g_4,g_5)$ is obtained as $\overline{\mathbf{g}} = \mathbf{q} \times \mathbf{g}$ where $\mathbf{q}$ is an arbitrary point on the line. Pl{\"u}cker coordinates fulfill the Pl{\"u}cker condition $\mathbf{g}^T\cdot\overline{\mathbf{g}} =0$. More on Pl\"ucker coordinates can be found in Pottmann and Wallner \cite[Chapter~2]{Pottmann2001}.

Study parameters or dual quaternions are used for the description of rigid body transformations either as 8-tuples $\mathbf{p} = (p_0, p_1, \ldots, p_7)$ or by using two quaternions $\mathbf{a}, \mathbf{b}$ as dual quaternion $\mathbf{p} = \mathbf{a} + \varepsilon \mathbf{b}$, where $\mathbf{a} = (p_0, p_1, p_2, p_3)$, $\mathbf{b} = (p_4,p_5,p_6,p_7)$ and $\varepsilon$ is the dual unit $\varepsilon^2=0$ (see Bottema and Roth \cite[Chapter~XIII]{bottema1979}). Study parameters have to fulfill the quadratic equation $\mathbf{a}^T \cdot \mathbf{b} = 0$, which is the equation of the so-called Study quadric. Points $\mathbf{q}$ and lines $\mathbf{l}$ can be embedded into the algebra of dual quaternions by $\mathbf{q} = (1,0,0,0,0,q_x,q_y,q_z)$, where $(q_x,q_y,q_z)$ are the coordinates of the point in the Euclidean space $E^3$ and $\mathbf{l}=(0, g_0, g_1, g_2, 0, g_3, g_4, g_5)$ where $g_i$ are the Pl{\"u}cker coordinates of the line. Transformation $\mathbf{p}$ transforms a point or a line by
\begin{equation}
    \mathbf{p} \mapsto \frac{\mathbf{p}_\varepsilon \mathbf{q} \, \mathbf{p}^*}{\mathbf{p}\mathbf{p}^*} \qquad \text{and} \qquad 
    \mathbf{l} \mapsto \frac{\mathbf{p}_\varepsilon \mathbf{l} \, \mathbf{p}_\varepsilon^*}{\mathbf{p}\mathbf{p}^*} ,
\end{equation}
where $\mathbf{p}_\varepsilon = \mathbf{a} - \varepsilon\mathbf{b}$ and the star denotes the conjugate of a dual quaternion.


A single point $\mathbf{p}$ on the Study quadric describes a discrete transformation. A curve $C(t)$ on the Study quadric describes a 1-parametric rigid body motion in SE(3). If all point trajectories are rational curves, $C(t)$ is called a rational motion. It is given by a polynomial $C(t)$ with dual quaternion coefficients.

%
%
\vspace{10pt}\noindent
\textbf{Synthesis of Rational Mechanisms -- Rational Motion Factorization}
\vspace{5pt}

\noindent
Given a rigid body motion represented by the polynomial $C(t)$ over the dual quaternions, it is generically possible to synthesize a linkage that performs this motion by combining different factorizations of $C(t)$, c.f. Heged\"us, Schicho, and Schr\"ocker \cite{Hegeds2013}.

We compute the factorizations of the motion polynomial with Python using the open source package \texttt{biquaternion-py} by Thimm \cite{bq-link-docs}.
This package provides a user-friendly implementation of a general dual quaternion algebra and polynomials over it for numerical and symbolical computations.
It is ready for use, even by non-specialists but some attention needs to be paid to details of symbolic computations (providing extension fields over which polynomials factor if they cannot be inferred implicitly) and numerics (for example taking into account rounding errors in polynomial division). More about the general usage of this package and its intricacies can be found in the documentation of the \texttt{biquaternion-py} package \cite{bq-link-docs}.

\section{Rational Mechanism Toolbox}\label{sec:toolbox}

While the previous section introduced methods and tools that are implemented but are already known, this section presents the main contribution of this work. It is the added functionality in terms of rapid prototyping of the linkages. Installation instructions can be found in the documentation hosted on the ReadTheDocs platform and are accessible at \cite{rat-link-docs}. The source is available from Gitlab repository hosted by the University of Innsbruck \cite{rat-link-gitlab}, which also serves as the code maintenance server.

As an example to demonstrate the features, we will use a general Bennett mechanism that was synthesized by Brunnthaler et al. \cite{brunnthaler2005new}. The motion curve of degree two of the given Bennett linkage has the equation specified by Study parameters
\begin{equation}
    C(t) = 
    \begin{bmatrix}
        0 \\
        22134 + 39870 t + 4440 t^2 \\
        -42966+9927t+16428 t^2 \\
        -115878-73843t-37296 t^2 \\
        0 \\
        -7812-14586t-1332 t^2 \\
        6510-1473t-2664 t^2 \\
        -3906-1881t-1332 t^2 \\
    \end{bmatrix}
\end{equation}

\begin{figure}[b]
    \centering
    \begin{subfigure}[t]{0.32\textwidth}
        \centering
        \includegraphics[width=\textwidth]{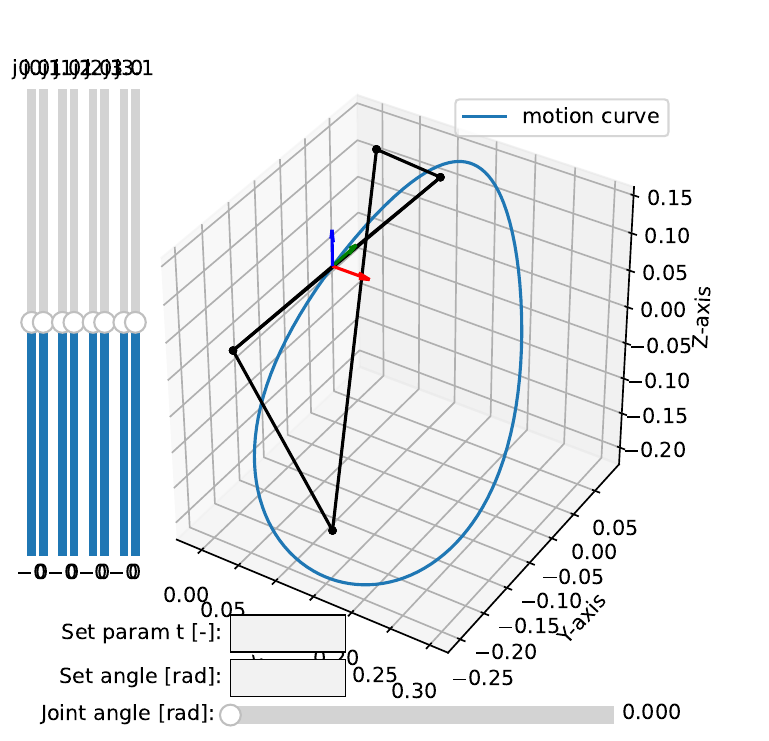}
    \end{subfigure}
    \hfill
    \begin{subfigure}[t]{0.32\textwidth}
        \centering
        \includegraphics[width=\textwidth]{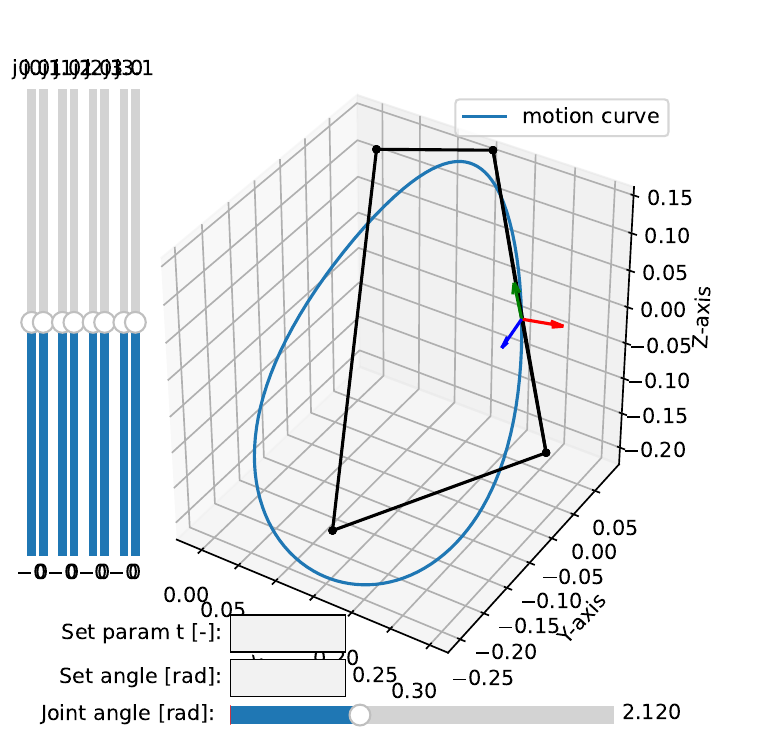}
    \end{subfigure}
    \hfill
    \begin{subfigure}[t]{0.32\textwidth}
        \centering
        \includegraphics[width=\textwidth]{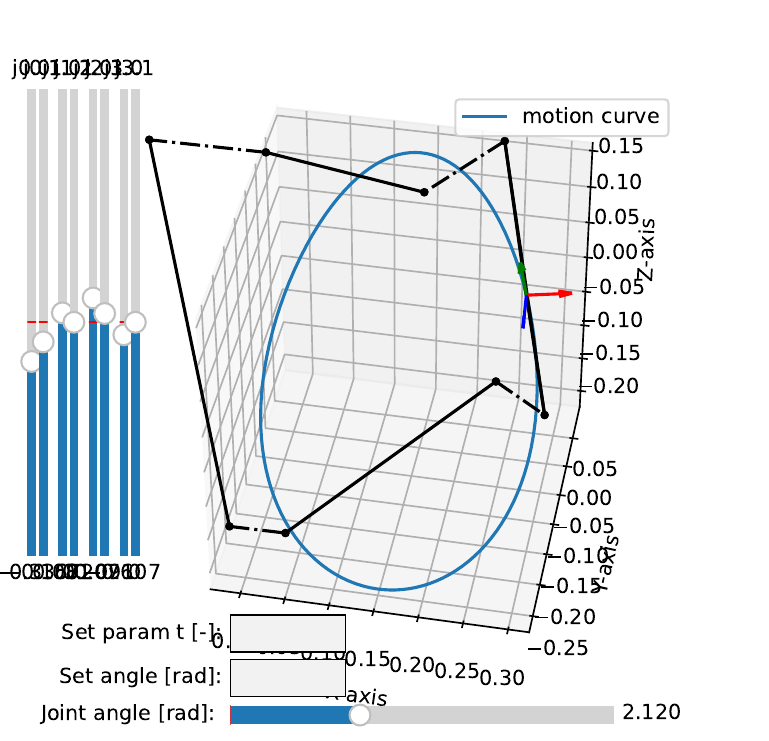}
    \end{subfigure}
    \caption{Bennett mechanism (black) with its tool frame and motion curve path (blue): in home configuration (left); moved to 2.120 rad (121.5 deg) of rotation of the driving joint axis (center); with designed joint segments -- dash-dotted lines (right).}
    \label{fig:plotting}
\end{figure}

This motion polynomial fulfills all properties needed to represent a rigid body motion on the Study quadric and it is also factorizable. Therefore, we can use the toolbox as follows in Algorithm \ref{alg:plotting} to plot the Bennett mechanism. The four lines of code in Algorithm \ref{alg:plotting} are capable of analyzing the motion curve $C(t)$ -- the only input, factorize it, and use the output representation defined in the Rational Mechanisms toolbox to visualize it interactively, as shown in Figure \ref{fig:plotting}.

\begin{algorithm}
\caption{Plotting a mechanism}\label{alg:plotting}
\begin{algorithmic}
\Require C(t)
\State c = RationalCurve(C(t)) \Comment{{\normalfont create an object Rational Curve from the input equations}}
\State factors = FactorizationProvider.factorize\textunderscore motion\textunderscore curve(c) \Comment{{\normalfont factorize curve}}
\State m = RationalMechanism(factors) \Comment{{\normalfont create a mechanism object}}
\State plt = Plotter.plot$(m)$ \Comment{{\normalfont plot the mechanism}}
\end{algorithmic}
\end{algorithm}

The interactive plot applies dual quaternion actions on points to animate the mechanism. The curve parameter $t$ is mapped to the angle of the driving joint from $0$ to $2 \pi$. In the background, the mechanism has not only its kinematic model but also a physical model that can be controlled via sliders next to the plot. The line model in Figure \ref{alg:plotting} (left, center) does not provide enough space to design joints for manufacturing. Nevertheless, the sliders can be used to manually obtain a line model whose realization will be free of self-collisions, as shown in Figure \ref{fig:plotting} (right).


Once a user is satisfied with the design, it may be desirable to save the mechanism in the current state. This can be done in the figure window by filling \emph{Save with filename} field and confirming by Enter, or using the command \texttt{RationalMechanism.save(filename)} in the Python console. A mechanism object can be recovered from a file using the command \texttt{ m = RationalMechanism .from\textunderscore saved\textunderscore file(filename)}, as shown in the examples in the documentation. 

In the toolbox, a collision check for the line model is implemented. It uses dual quaternion operations on lines to create parametric equations of moving lines, joints and links, during a full cycle of the motion. The lines move relatively between each other and two lines $\mathbf{l_0}$ and $\mathbf{l_1}$ represented by their Pl{\"u}cker coordinates intersect if the following equation is fulfilled \cite[p.140]{Pottmann2001}:
\begin{equation}
    \mathbf{g_0} \cdot \overline{\mathbf{g_1}} + \overline{\mathbf{g_0}} \cdot \mathbf{g_1} = 0
\end{equation}
After the equations are solved, the physical link or joint line-segments of $\mathbf{l_0}$ and $\mathbf{l_1}$ are checked if it is a true linkage-physical-model collision, or the lines collide somewhere out of the physical realization. For all joint-joint, joint-link, and link-link collision scenarios, tens of intersections can occur in space for a 4R or 6R with one DoF. Additionally, some of the relative paths are given by polynomials of higher degree. Due to these two reasons and Python specifications, the collision check may take from seconds to a low amount of minutes, depending on the available computational power. The collision check is also implemented to run in parallel on multi-core computers, which in general provides the results much faster. The collision analysis may be called using \texttt{RationalMechanism.collision\textunderscore check()} method.



\vspace{10pt}\noindent
\textbf{Generation of Design Parameters and CAD Model Prototyping}
\vspace{5pt}

\noindent
In the package, the command \texttt{RationalMechanism.get\textunderscore design(scale)} generates design parameters for a pre-prepared CAD model of a link that is available on Onshape -- an online CAD system platform. Anyone can access, edit, export, and download it. For more details and access, see \cite{bennett-zenodo}, where the example Bennett mechanism files, ready for 3D printing, were also uploaded. The method \texttt{get\textunderscore design} returns a tuple of Denavit-Hartenberg (DH) parameters, with proper automatic placement \cite{Huczala2022iccma} of coordinate frames following the standard DH convention, and a set of Connection Point-parameter pairs for every link. The length parameters of the Bennett linkage from Figure \ref{fig:plotting} (right), scaled by 200 [-], result in the output of Table \ref{tab:dhcp}.

\begin{table}[h]
    \centering
    \begin{tabular}{c|c|c|c|c|c}
        $i$ & $d_i$ [mm] & $a_i$ [mm] & $\alpha_i$ [deg] & $cp_{0i}$ [mm] & $cp_{1i}$ [mm] \\ \hline
        0 & 64.580219 & 48.517961 & -144.679172 & 2.085621 & 17.491631 \\ 
        1 & 0 & 83.708761 & -94.053746 & -3.508369 & -0.650840 \\
        2 & 0 & 48.517961 & -144.679172 & -21.650840 & 39.381058 \\
        3 & 0 & 83.708761 & -94.053746 & 60.381058 & -83.494598 \\ 
    \end{tabular}
    \caption{DH parameters and Connecting Points of the Bennett linkage}
    \label{tab:dhcp}
\end{table}

\begin{figure}[t]
    \centering
    \begin{subfigure}[t]{0.4\textwidth}
        \centering
        \includegraphics[width=\textwidth]{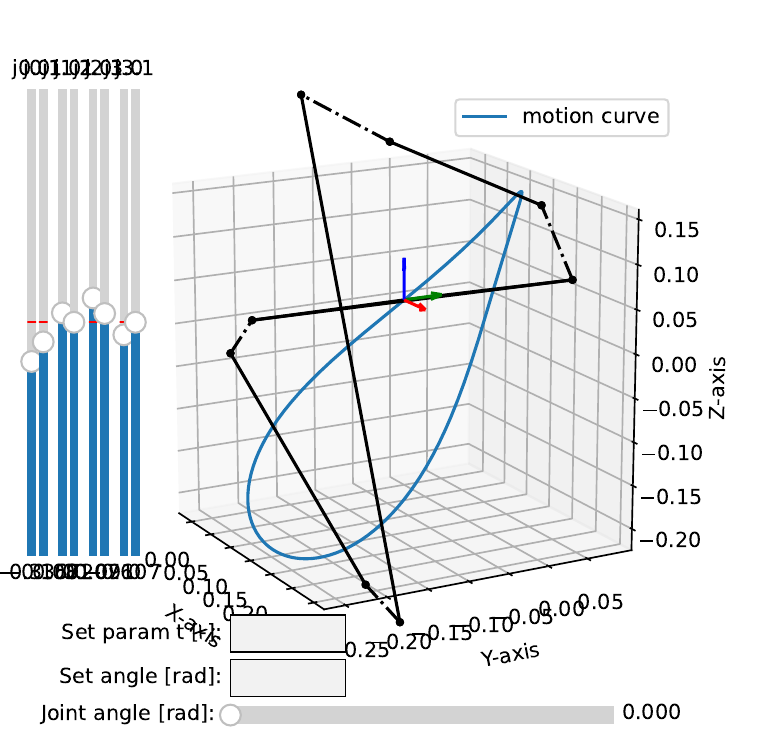}
    \end{subfigure}
    \hfill
    \begin{subfigure}[t]{0.25\textwidth}
        \centering
        \includegraphics[width=\textwidth]{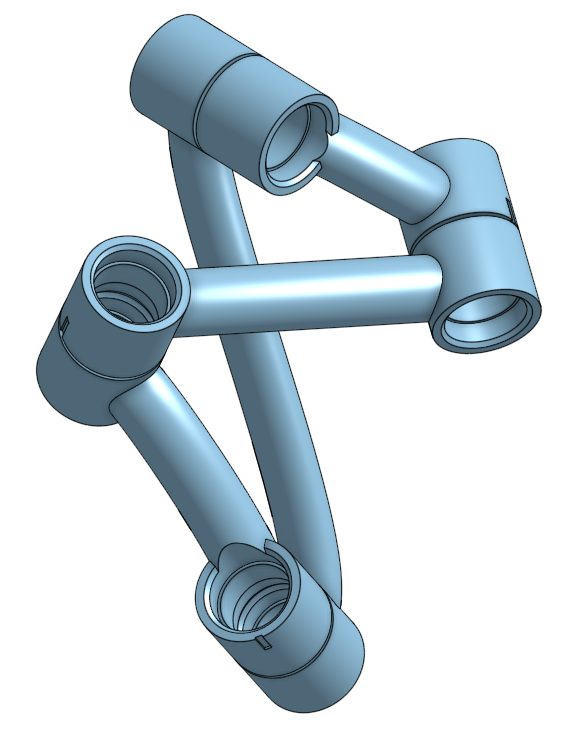}
    \end{subfigure}
    \hfill
    \begin{subfigure}[t]{0.25\textwidth}
        \centering
        \includegraphics[width=\textwidth]{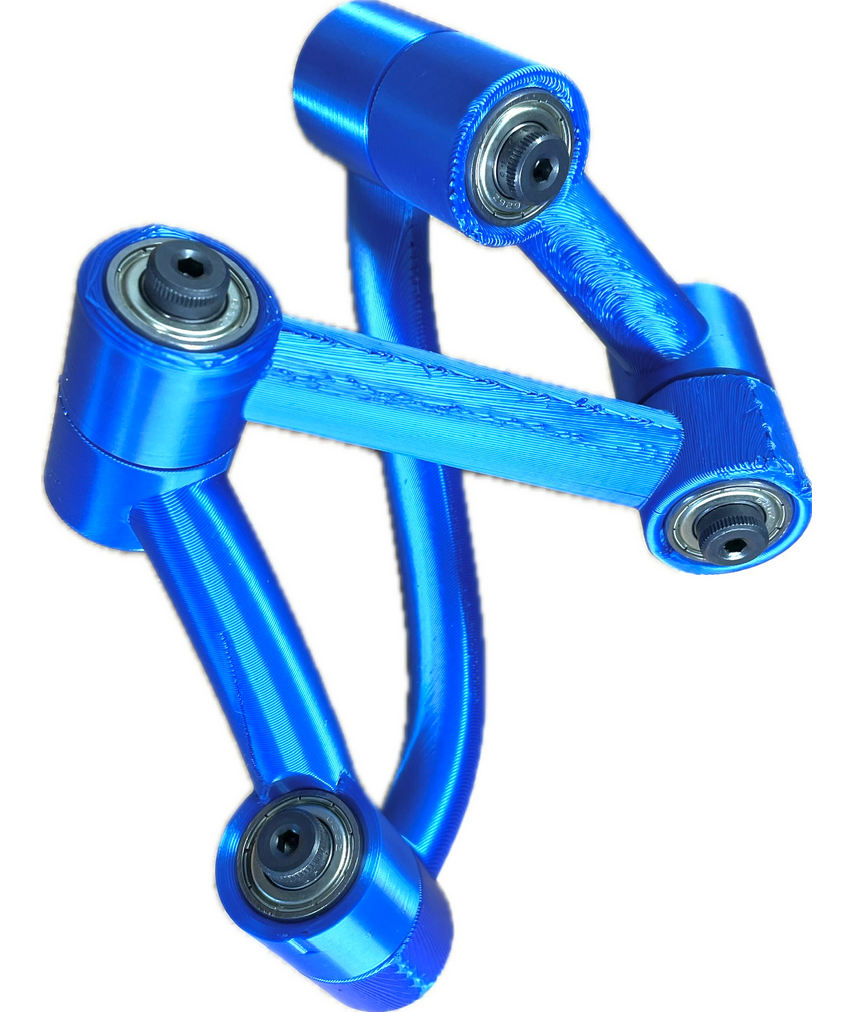}
    \end{subfigure}
    
    \caption{Bennett mechanism: line model (left); CAD model (center); 3D-printed and assembled collision-free prototype (right).}
    \label{fig:bennet-real}
\end{figure}

The scaling allows one to model a physical linkage with the exact values from the table. The numerical values from Table \ref{tab:dhcp} were inserted into the pre-prepared CAD model, 3D printed, assembled, and the result may be seen in Figure \ref{fig:bennet-real}. By default, the joint-segment of the CAD model is 41 mm long, and the $cp$ parameters are mapped to fit this length if the line model in the visualization tool is not matching it already. This length can be optionally changed. 

Note that the joint angles among the DH are irrelevant in this procedure as every joint undergoes a full-cycle motion. 
Also, we want to point out that the parameter $d_0$, the distance from the base coordinate frame along the z-axis, is truly non-zero. Setting it to zero would make the values of connection points $cp_{00}$ and $cp_{13}$ invalid.

\newpage

\vspace{10pt}\noindent
\textbf{Motion Interpolation}
\vspace{5pt}

\noindent
As mentioned in the introduction, the 4 poses interpolation method from \cite{Hegeds2015} is implemented in the package to synthesize 6R linkages with 1 DoF. The technique is based on computing a cubic motion polynomial interpolating the given poses. However, as explained in the paper, such a polynomial need not always to exist. Additionally, the authors impose restrictions on the given poses to avoid planar or spherical mechanisms and prismatic joints. We refer to \cite{Hegeds2015} for potential pitfalls in case the interpolation in the Rational Linkages package does not yield satisfactory results.

\begin{figure}[t]
    \centering
    \begin{subfigure}[t]{0.48\textwidth}
        \centering
        \includegraphics[trim={4.5cm 2cm 0 0}, clip, width=\textwidth]{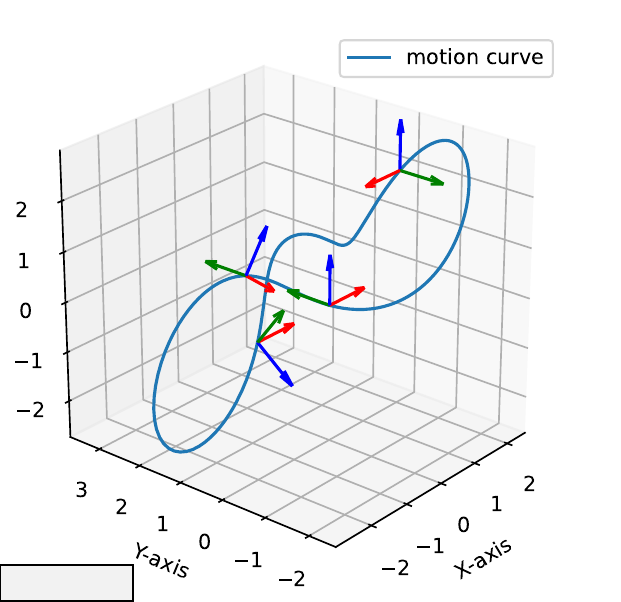}
    \end{subfigure}
    \hfill
    \begin{subfigure}[t]{0.48\textwidth}
        \centering
        \includegraphics[trim={4.8cm 2cm 0 0}, clip, width=\textwidth]{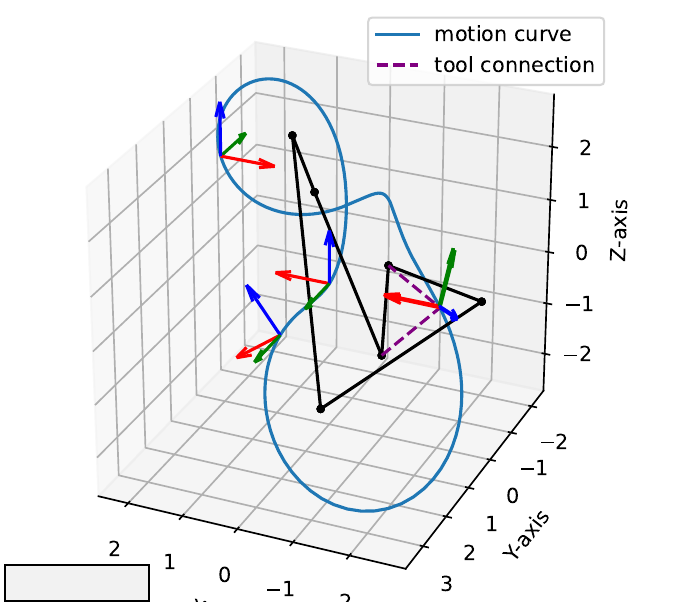}
    \end{subfigure}
    
    \caption{Rational motion interpolation: 4 given poses with interpolated curve (left); mechanism in $\mathbf{p}_2$ (right).}
    \label{fig:interp}
\end{figure}

A motion interpolation example will be shown for four poses comprising the identity $\mathbf{p}_0=[1,0,0,0,0,0,0,0]^T$ and the three poses $\mathbf{p}_{1}$, $\mathbf{p}_2$, $\mathbf{p}_3$ defined by Study parameters:
\begin{equation}  
    \mathbf{p}_1 = [0, 0, 0, 1, 1, 0, 1, 0]^T\!\!,\    
    \mathbf{p}_2 = [1, 2, 0, 0, -2, 1, 0, 0]^T\!\!,\  
    \mathbf{p}_3 = [3, 0, 1, 0, 1, 0, -3, 0]^T\!\!.
\end{equation}
The rational curve $C(t)$ that interpolates these poses can be obtained using the command \texttt{MotionInterpolation.interpolate([p0, p1, p2, p3])} from the toolbox. The obtained equation is of degree three and has the form

\begin{equation}
    C(t) = 
    \begin{bmatrix}
        t^3 - 0.4375t^2 - 0.171875t, \\
        0.25t^2 - 0.25t - 0.078125, \\
        0.3125t^2 - 0.078125t - 0.0390625, \\
        -0.0625t^2 + 0.109375t - 0.0390625, \\
       6 0.28125t, \\
        0.125t^2 - 0.125t - 0.0390625, \\
        -t^2 + 0.34375t + 0.078125, \\
        0 \\
    \end{bmatrix}
\end{equation}

which corresponds to an overconstrained 6R mechanism. The polynomial can be factorized and the corresponding 6R linkage is visualized in Figure~\ref{fig:interp}.

\section{Conclusion}

This paper briefly presents a new Python-based package that implements various algorithms to deal with rational single-loop linkage synthesis and design, with the intention of simplifying their prototyping. Ready-to-run Python scripts and extended information on the presented examples, including results that were too long to present here, may be found in the documentation page \cite{rat-link-docs} specially created for this paper. 

Future development will be focused on collision-free realizations of these linkages, which is a topic that has not been much investigated. The algorithm by Li, Nawratil, et al. \cite{Li2020} which almost always finds a collision-free line model of any given 6R mechanism is also planned to be implemented. Furthermore, curved links can be applied, as in \cite{Kot2020, Mlotek2023}, which can greatly expand the design possibilities, or the connection to a multibody dynamics system, as Exudyn \cite{Gerstmayr2023}, may be developed. This will bring closed-loop linkages much closer to engineering applications, serving, for example, as cheap single-purpose devices in robotic manipulation.

\vspace{15pt}
\noindent
\textbf{Acknowledgements}
\vspace{5pt}

\small
\begin{wrapfigure}{l}{0.31\textwidth}
    \centering
    \vspace{-18pt}
    \includegraphics[width=0.34\textwidth]{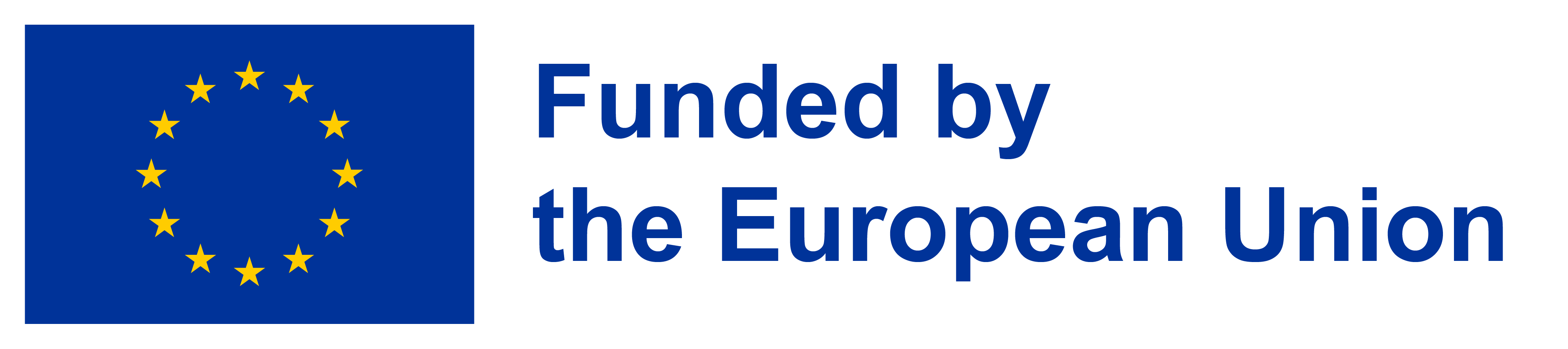}
    \vspace{-35pt}
\end{wrapfigure}
\scriptsize
\noindent
\textbf{Disclaimer} Funded by the European Union. Views and opinions expressed are however those of the author(s) only and do not necessarily reflect those of the European Union or the European Research Executive Agency (REA). Neither the European Union nor the granting authority can be held responsible for them.

\vspace{10pt}
\noindent Johannes Siegele was supported by the Austrian Science Fund (FWF): P 31888

\printbibliography

@article{Huczala2022,
  author={Huczala, D. and Kot, T. and Pfurner, M. and Krys, V. and Bobovský, Z.},
  journal={IEEE Access}, 
  title={Multirepresentations and Multiconstraints Approach to the Numerical Synthesis of Serial Kinematic Structures of Manipulators}, 
  year={2022},
  volume={10},
  pages={68937-68951},
  doi={10.1109/ACCESS.2022.3186098}}

@article{Dogra2022,
  title = {Optimal Synthesis of Unconventional Links for Modular Reconfigurable Manipulators},
  ISSN = {1528-9001},
  url = {http://dx.doi.org/10.1115/1.4054336},
  DOI = {10.1115/1.4054336},
  journal = {Journal of Mechanical Design},
  publisher = {ASME International},
  author = {Dogra,  A. and Sekhar Padhee,  S. and Singla,  E.},
  year = {2022}
}

@article{Kot2020,
  title = {Finding Optimal Manipulator Arm Shapes to Avoid Collisions in a Static Environment},
  ISSN = {2076-3417},
  url = {http://dx.doi.org/10.3390/app11010064},
  DOI = {10.3390/app11010064},
  journal = {Applied Sciences},
  publisher = {MDPI AG},
  author = {Kot,  T. and Bobovský,  Z. and Brandst\"{o}tter,  M. and Krys,  V. and Virgala,  I. and Novák,  P.},
  year = {2020}
}

@article{Xi2018,
  title = {Type Synthesis of Walking Robot Legs},
  volume = {31},
  ISSN = {2192-8258},
  url = {http://dx.doi.org/10.1186/s10033-018-0216-7},
  DOI = {10.1186/s10033-018-0216-7},
  number = {1},
  journal = {Chinese Journal of Mechanical Engineering},
  publisher = {Springer Science and Business Media LLC},
  author = {Xi,  D. and Gao,  F.},
  year = {2018},
}

@article{Zhang2019,
  title = {Kinematic synthesis method for the one-degree-of-freedom jumping leg mechanism of a locust-inspired robot},
  volume = {63},
  ISSN = {1869-1900},
  url = {http://dx.doi.org/10.1007/s11431-019-9750-6},
  DOI = {10.1007/s11431-019-9750-6},
  number = {3},
  journal = {Science China Technological Sciences},
  publisher = {Springer Science and Business Media LLC},
  author = {Zhang,  Z. and Yang,  Q. and Zhao,  J. and Chang,  B. and Liu,  X.},
  year = {2019},
  pages = {472–487}
}

@inbook{Hegeds2012,
  title = {Construction of Overconstrained Linkages by Factorization of Rational Motions},
  ISBN = {9789400746206},
  url = {http://dx.doi.org/10.1007/978-94-007-4620-6_27},
  DOI = {10.1007/978-94-007-4620-6_27},
  booktitle = {Latest Advances in Robot Kinematics},
  publisher = {Springer Netherlands},
  author = {Heged\"{u}s,  G. and Schicho,  J. and Schr\"{o}cker,  H.-P.},
  year = {2012},
  pages = {213–220}
}

@article{Hegeds2013,
  title = {Factorization of rational curves in the {Study} quadric},
  volume = {69},
  ISSN = {0094-114X},
  url = {http://dx.doi.org/10.1016/j.mechmachtheory.2013.05.010},
  DOI = {10.1016/j.mechmachtheory.2013.05.010},
  journal = {Mechanism and Machine Theory},
  publisher = {Elsevier BV},
  author = {Heged\"{u}s,  G. and Schicho,  J. and Schr\"{o}cker,  H.-P.},
  year = {2013},
  pages = {142–152}
}

@article{Hegeds2015,
  title = {Four-Pose Synthesis of Angle-Symmetric {6R} Linkages},
  volume = {7},
  ISSN = {1942-4310},
  url = {http://dx.doi.org/10.1115/1.4029186},
  DOI = {10.1115/1.4029186},
  number = {4},
  journal = {Journal of Mechanisms and Robotics},
  publisher = {ASME International},
  author = {Heged\"{u}s,  G. and Schicho,  J. and Schr\"{o}cker,  H.-P.},
  year = {2015},
}

@article{Li2020,
  title = {Invertible Paradoxic Loop Structures for Transformable Design},
  volume = {39},
  ISSN = {1467-8659},
  url = {http://dx.doi.org/10.1111/cgf.13928},
  DOI = {10.1111/cgf.13928},
  number = {2},
  journal = {Computer Graphics Forum},
  publisher = {Wiley},
  author = {Li,  Z. and Nawratil,  G. and Rist,  F. and Hensel,  M.},
  year = {2020},
  pages = {261–275}
}

@article{Yihun2014,
  title = {Link-Based Performance Optimization of Spatial Mechanisms},
  volume = {136},
  ISSN = {1528-9001},
  url = {http://dx.doi.org/10.1115/1.4028304},
  DOI = {10.1115/1.4028304},
  number = {12},
  journal = {Journal of Mechanical Design},
  publisher = {ASME International},
  author = {Yihun,  Y. and Bosworth,  K. W. and Perez-Gracia,  A.},
  year = {2014}
}

@inbook{Mlotek2023,
  title = {The Effect of Deformation on Robot Shape - Changing Link},
  ISBN = {9783031326066},
  ISSN = {2211-0992},
  url = {http://dx.doi.org/10.1007/978-3-031-32606-6_54},
  DOI = {10.1007/978-3-031-32606-6_54},
  booktitle = {Mechanisms and Machine Science},
  publisher = {Springer Nature Switzerland},
  author = {Mlotek,  J. and Suder,  J. and Vocetka,  M. and Bobovský,  Z. and Krys,  V.},
  year = {2023},
  pages = {461}
}

@misc{bq-link-docs,
  author = {D. Thimm},
  title = {Biquaternion-py -- Documentation},
  note = {[Online] Accessed: 2024-01-15},
  howpublished = {\url{https://biquaternion-py.readthedocs.io}}  
}

@misc{rat-link-gitlab,
  author = {D. Huczala},
  title = {Rational {L}inkages -- source code repository},
  note = {GitLab instance of the University of Innsbruck (UIBK). [Online] Accessed: 2024-01-15},
  howpublished = {\url{https://git.uibk.ac.at/geometrie-vermessung/rational-linkages}}  
}

@misc{rat-link-docs,
  author = {D. Huczala},
  title = {Rational {L}inkages -- Documentation},
  note = {[Online] Accessed: 2024-01-15},
  howpublished = {\url{https://rational-linkages.readthedocs.io/latest/tutorials/ark2024.html}}  
}

@article{brunnthaler2005new,
  title={A new method for the synthesis of {B}ennett mechanisms},
  author={Brunnthaler, K. and Schr{\"o}cker, H.-P. and Husty, M.},
  year={2005},
  journal = {International Workshop on Computational Kinematics}
}

@book{Pottmann2001,
  title = {Computational Line Geometry},
  ISBN = {9783642040184},
  ISSN = {1612-3786},
  url = {http://dx.doi.org/10.1007/978-3-642-04018-4},
  DOI = {10.1007/978-3-642-04018-4},
  journal = {Mathematics and Visualization},
  publisher = {Springer},
  author = {Pottmann,  H. and Wallner,  J.},
  year = {2001}
}

@misc{bennett-zenodo,
  author = {D. Huczala},
  title = {Rational {L}inkages -- example of {B}ennett mechanism ({CAD} models weblinks and {STLs})},
  note = {DOI: 10.5281/zenodo.10479379},
  year = {2024},
  publisher = {Zenodo},
  howpublished = {\url{https://zenodo.org/doi/10.5281/zenodo.10479379}}  
}

@inproceedings{Huczala2022iccma,
  title = {An Automated Conversion Between Selected Robot Kinematic Representations},
  url = {http://dx.doi.org/10.1109/ICCMA56665.2022.10011595},
  DOI = {10.1109/iccma56665.2022.10011595},
  booktitle = {2022 10th International Conference on Control,  Mechatronics and Automation (ICCMA)},
  publisher = {IEEE},
  author = {Huczala,  D. and Kot,  T. and Mlotek,  J. and Suder,  J. and Pfurner,  M.},
  year = {2022}
}

@book{bottema1979,
  title={Theoretical Kinematics},
  author={Bottema, O. and Roth, B.},
  publisher={North-Holland Publishing Company},
  year={1979}
}

@article{Gerstmayr2023,
  title = {Exudyn – a {C}++-based {P}ython package for flexible multibody systems},
  ISSN = {1573-272X},
  url = {http://dx.doi.org/10.1007/s11044-023-09937-1},
  DOI = {10.1007/s11044-023-09937-1},
  journal = {Multibody System Dynamics},
  publisher = {Springer Science and Business Media LLC},
  author = {Gerstmayr,  Johannes},
  year = {2023},
  month = oct 
}







\end{document}